\documentclass{article}
\usepackage[utf8]{inputenc}
\usepackage{setspace}
\usepackage[a4paper, total={6in, 10in}]{geometry}
\usepackage[T1]{fontenc}
\usepackage{graphicx}
\usepackage{natbib}
\usepackage{authblk}
\RequirePackage{natbib}
\usepackage{amssymb,amsfonts,amsmath,amsthm,amscd,dsfont,mathrsfs,mathtools,nicefrac}
\usepackage{float,psfrag,epsfig,color,xcolor,url,hyperref}
\usepackage{epstopdf,bbm,mathtools,enumitem}
\usepackage{url}

\def\E{\mathbb{E}} % Expectation symbol

\def\balign#1\ealign{\begin{align}#1\end{align}}
\def\baligns#1\ealigns{\begin{align*}#1\end{align*}}
\def\balignat#1\ealign{\begin{alignat}#1\end{alignat}}
\def\balignats#1\ealigns{\begin{alignat*}#1\end{alignat*}}
\def\bitemize#1\eitemize{\begin{itemize}#1\end{itemize}}
\def\benumerate#1\eenumerate{\begin{enumerate}#1\end{enumerate}}

\newenvironment{talign*}
 {\csname align*\endcsname}
 {\endalign}
\newenvironment{talign}
 {\csname align\endcsname}
 {\endalign}

\def\balignst#1\ealignst{\begin{talign*}#1\end{talign*}}
\def\balignt#1\ealignt{\begin{talign}#1\end{talign}}
\let\originalleft\left
\let\originalright\right
\renewcommand{\left}{\mathopen{}\mathclose\bgroup\originalleft}
\renewcommand{\right}{\aftergroup\egroup\originalright}

\makeatletter
\let\original@algocf@latexcaption\algocf@latexcaption
\long\def\algocf@latexcaption#1[#2]{%
  \@ifundefined{NR@gettitle}{%
    \def\@currentlabelname{#2}%
  }{%
    \NR@gettitle{#2}%
  }%
  \original@algocf@latexcaption{#1}[{#2}]%
}
\makeatother

\title{Challenges for Reinforcement Learning in Healthcare}
\begin{document}

\author[1,4]{Elsa Riachi \thanks{elsa.riachi@mail.utoronto.ca}}
\author[2,5]{Muhammad Mamdani \thanks{muhammad.mamdani@unityhealth.to}}
\author[3]{Michael Fralick \thanks{mike.fralick@mail.utoronto.ca}}
\author[1,4,5]{Frank Rudzicz \thanks{frank@cs.toronto.edu}}

\affil[1]{University of Toronto, Department of Computer Science}
\affil[2]{Li Ka Shing Centre for Healthcare Analytics Research and Training}
\affil[3]{Sinai Health System, Department of Medicine, University of Toronto}
\affil[4]{Vector Institute for Artificial Intelligence}
\affil[5]{Unity Health Toronto}

\maketitle

\begin{abstract}

Many healthcare decisions involve navigating through a multitude of treatment options in a sequential and iterative manner to find an optimal treatment pathway with the goal of an optimal patient outcome. Such optimization problems may be amenable to reinforcement learning. A reinforcement learning agent could be trained to provide treatment recommendations for physicians, acting as a decision support tool. However, a number of difficulties arise when using RL beyond benchmark environments, such as specifying the reward function, choosing an appropriate state representation and evaluating the learned policy. 
    
\end{abstract}

\section{Introduction}

Along with ongoing changes to regulatory approval processes for software \cite{elenko2015regulatory}, machine learning is being increasingly used within healthcare. Machine learning can be broadly divided into supervised learning, unsupervised learning, and reinforcement learning. Supervised learning requires a dataset where the outcome of interest is known, and results in a model which categorizes data points, e.g., as images of skin cancer or CT scans, by finding correlative or discriminative relationships in fully labelled data \cite{Esteva2017}. For example, supervised learning has been applied to identify hospitalized patients who are at an increased risk of death using collected data of vital signs and patient outcomes \cite{escobar2020automated}. Clinical data can also be leveraged to uncover hidden patterns with unsupervised learning. Unlike supervised learning which is used to obtain a predictive model, unsupervised learning can be used to better understand patient data. For example, unsupervised learning was recently leveraged to identify whether there might be distinct clinical phenotypes of patients hospitalized with sepsis \cite{seymour2019derivation}.  While predictions made by a supervised learning model can be used to decide the next course of action, supervised learning does not find the optimal sequence of actions directly, since the effect of an action taken at a given time step is not independent of subsequent actions. Reinforcement learning therefore lends itself well to problems of sequential decision making where the effect of actions may extend over an unknown time duration into the future.  Reinforcement learning, the focus of this article, attempts to make a sequence of decisions to achieve a given goal. Unlike supervised or unsupervised methods, it is used to identify the optimal sequence of actions based on their subsequent effect on the state (e.g., a patient’s current condition).  
In this survey, we highlight the key challenges that arise when using RL to learn improved treatment strategies from medical records, and point the reader towards salient directions for future research.

\section{Reinforcement learning}\label{sec:rl}

A reinforcement learning (RL) agent is trained to take optimal actions with respect to a chosen reward function and assumed state of the world. The reward signal acts as a \textit{reinforcer} of desired behaviour and represents the task we want the agent to perform. The agent learns a policy that maximizes cumulative reward by interacting with its environment. Formally, our goal is to solve a Markov decision process, represented by a tuple $(S, A, r, \gamma, P)$, where $S$ denotes the state space, $A$ the action space, $r$ the reward function, $\gamma$ the discount factor, and $P$ the environment's transition probabilities. The environment dynamics are assumed to satisfy the Markov property, i.e.:
\begin{equation}
    P(s_{t+1} | h_t) \approx P(s_{t+1} | s_t, a_t),
    \label{eq:markov}
\end{equation}
where $h_t = ((s_0, a_0), \ldots, (s_t, a_t))$ is the history of all state-action pairs up to and including time $t$.  The Markov assumption is not restrictive -- if the environment dynamics do not satisfy the Markov property for a given state representation, another state representation can be used to act as a summary of the environment's history from which the next state can be predicted. The current state can then be written as a function of past state-action pairs  $s_{t} = f((s_0, a_0)...(s_{t-1}, a_{t-1}))$ or $s_t = f(h_{t-1})$. The choice of state representation is discussed in Section \ref{sec:StateRep}. 
The goal of the RL agent is to maximize the expected return, or the discounted sum of rewards  $\sum_{t=0}^{T} \gamma^t r_t$. The agent's policy $\pi(a\,|\,s)$ assigns a probability to actions $a \in A$, conditioned on the current state $s$. The state value function of a policy $\pi$, denoted as $V^\pi(s)$, is the expected return obtained after being in state $s$ and subsequently acting according to policy $\pi$. The action value function of a policy $\pi$, denoted as $Q^\pi(a, s)$, is the expected return obtained after being in state $s$, taking action a and subsequently acting according to policy $\pi$. 

The Bellman Equations \ref{eq:bellman_v} and \ref{eq:bellman_q} define the state value and action-value functions recursively. 

\begin{align}
    V^\pi(s_{t}) &= r(s_t, a_t) + \gamma \E_{s_{t+1} \sim P(. | s, a)} V^\pi(s_{t+1}) \label{eq:bellman_v} \\
    Q^\pi (a_t, s_t) &= r(s_t, a_t) + \gamma \E_{a_{t+1} \sim \pi(. | s), s_{t+1} \sim P(. | s, a)} Q^\pi (a_{t+1}, s_{t+1})
    \label{eq:bellman_q}
\end{align}

It is clear that an optimal policy $\pi^*$ will select the action $a$ for which the action-value function $Q^{\pi^*}(a, s)$ is highest. The Bellman optimality Equations \ref{eq:bellman_v_opt} and \ref{eq:bellman_q_opt} use this fact to recursively define the value functions for an optimal policy $\pi^*$. 

\begin{align}
    V^{\pi*}(s_{t}) &= \max_{a} Q^{\pi^*}(a, s_t)
    \label{eq:bellman_v_opt} \\
    Q^{\pi^*} (a_t, s_t) &= r(s_t, a_t) + \gamma \E_{s _{t+1} \sim P(. | s_t, a_t)} \max_{a_{t+1}} Q^\pi (a_{t+1}, s_{t+1})
    \label{eq:bellman_q_opt}
\end{align}

If the environment dynamics, that is, the transition probabilities $P(s_{t+1} | s_t, a_t)$ are known, and the state and action spaces are finite, an optimal policy can be learned using an iterative procedure called policy iteration, which alternates between a policy evaluation step and a policy improvement step \cite{sutton2018reinforcement}. 

Generally, the transition dynamics are unknown. By interacting with the environment, value-based RL agents observe state transitions and rewards and use those observations to build an approximation of $Q^{\pi^*}(s, a)$ from which the optimal policy is obtained. Off-policy algorithms can learn an optimal policy from experience generated by a behavioural policy that is different from the current estimate of the optimal policy. $Q$-learning \cite{watkins1992} is an off-policy, value-based algorithm. In the tabular case, the action-value function $Q$ is updated according to Equation \ref{eq:q_tabular}. Although the transition tuples used for the updates are generated by a suboptimal behavioural policy, $Q$-learning converges to $Q^{\pi^*}$ if all the state-action pairs are visited infinitely often \cite{watkins1992}. 

\begin{equation}
    Q(s_t, a_t) = Q(s_t, a_t) + \alpha(r(s_t, a_t) + \gamma \max_a Q(s_{t+1}, a) - Q(s_t, a_t))\
    \label{eq:q_tabular}
\end{equation}

In the case of a large or continuous action space or state space, a function approximator can be used to represent the $Q$-function termed the critic. For instance, the Q-function can be learned using kernel-based regression methods \cite{fitted_q} or a neural network \cite{neural_q}.  Q-learning with a function approximator is termed Fitted Q-Iteration (FQI) to distinguish it from Q-learning in which state-action values are stored in a tabular structure.  The parameters $\theta $ of the critic can be updated using Equation \ref{eq:update}. In practice the expectation over state transitions is estimated using a sample average from a batch of recorded trajectories.

\begin{equation}
    \theta =  \theta + \alpha \mathbb{E}\,[(r(s_t, a_t) + \gamma V(s_{t+1}) - Q(s_t, a_t))\nabla_\theta Q(s_t, a_t)\,]
    \label{eq:update}
\end{equation}

The policy can also be parameterized and updated to increase its return according to an advantage term $Q^{\pi}(S_t, A_t) - V^{\pi}(S_t)$, which is computed using the estimated critic. Alternating a policy improvement step with a policy evaluation step using gradient updates constitutes an actor-critic algorithm \cite{konda2000actor}, and is seen as a generalization of policy iteration. 

In the observational setting, the agent is considered to interact with the environment by sampling experiences from previously collected records. Many applications of RL in healthcare \cite{raghu2017continuous,raghu2018,optimalheparin,Weng2017} fall under this category because it is infeasible or even unethical to evaluate a policy on patients. Often, we would like to discover better policies than those reflected in the records. For example, Raghu {\em et al.} ~\cite{raghu2017continuous} used data from the ICU to learn better treatment strategies for managing sepsis. However, this requires evaluating a different policy in the evaluation step than the one used to generate the recorded experience. In this case, we resort to off-policy reinforcement learning algorithms. If the historical data is used to build a model of the environment, as was done by Raghu {\em et al.} ~\cite{raghu2017continuous}, the model could be used to simulate trajectories using the policy network. The critic, or $Q$-function, can then be updated using trajectories generated by the current estimate of the optimal policy. 
%\fi

\section{Reinforcement learning from observational data}\label{sec:rl_obs}
% reiterating the difference between rl and supervised ml 
Reinforcement learning leverages past experiences which involve interacting with the environment. Unlike supervised learning, where labels specify the required or desired output, past experience used for training reinforcement learning algorithms may be suboptimal. Trajectories consist of sequences of state-action pairs which may result in an undesired outcome. In the case of dynamic, personalized treatment recommendation systems, the data may consist of patient trajectories recorded in electronic health records (EHRs), which include information recorded over a patient's course of treatment, such as vital signs, lab measurements, administered medication or interventions.  Sequential data in EHRs is used to construct state-action pairs which are subsequently used to train a reinforcement learning algorithm.  The key difference between RL using observational data and RL using simulated environments is the lack of exploration during training in the former. Much of the success of reinforcement learning has been in simulated robotics \cite{DBLP:journals/corr/AndrychowiczWRS17} or games where exploration is fast and unconstrained \cite{Mnih2015HumanlevelCT, silver2017mastering}. Exploration is easy in a simulated environment, but is not always feasible or safe in the healthcare setting. Off-policy reinforcement learning algorithms rely on having good estimates of action-value functions, which require extensive samples of the state and action spaces. An exploration strategy is usually used to gather experience, such as occasionally taking a random action instead of the supposed optimal one. Batch reinforcement learning refers to learning a policy only from previously collected experiences. Fujimoto {\em et al.} ~\cite{fujimoto2018} showed how batch RL algorithms are susceptible to extrapolation errors, where the $Q$-function or critic overestimates the expected return for unseen state-action pairs, even when the training set consists of trajectories generated by an optimal policy. Observational data such as medical records consist of limited coverage or limited samples of the possible state-action pairs. The observational data is also collected under different policies than the one we wish to evaluate. Updates to the parameters of the $Q$-function occur under a different distribution than the on-policy distribution. As a result the learned policy may be optimal for a biased estimate of the $Q$-function. This can also lead to misleading evaluations when the learned policy is compared to that of the clinicians'. For example, Raghu {\em et al.}~\cite{raghu2017deep} used a reward function for treating sepsis that included the SOFA score, which is a clinical score that quantifies the degree of organ failure \cite{vincent1996sofa}. Their reward function penalized high SOFA scores as well as increases in SOFA scores. The learned policy recommended that patients with severe sepsis (and, hence, high SOFA scores) receive less aggressive treatment. The authors reasoned that this is an effect of having fewer severe cases in the dataset. However, Gottesman {\em et al.}~ \cite{gottesman2018evaluating} remarked that during training, the agent experienced trajectories where severe sepsis is aggressively treated with IV fluids and vasopressors, however its observed reward is diminished because high SOFA scores, which are correlated with severe cases, are penalized. The agent cannot learn that lack of treatment might result in lower returns because such trajectories are not present in the collected records. 

Evaluating a policy using experience collected under a different policy is termed `off-policy evaluation'. The policy that generated the collected experience is called the `behavioural policy', and the policy we wish to evaluate is called the `target' or `evaluation policy'. The goal is to estimate the expected total reward that an agent would obtain if it were acting under the evaluation policy. One approach is to use the importance sampling (IS) estimator shown in Equation \ref{eq:IS}, which is unbiased, but can suffer from high variance if the target and evaluation policies do not have a large common support \cite{Precup2000Eligibility}. The IS estimator computes the expected return of the evaluation policy $\pi^{e}$ using trajectories $\tau$ generated by a behavioural policy $\pi^{b}$, by reweighing the return of each observed trajectory $G(\tau)$ with its importance weight $\rho(\tau)$ shown in Equation \ref{eq:imp_weights}. 

\begin{equation}
    V^{\pi_e}  = \sum_{\tau} \rho(\tau) G(\tau)
    \label{eq:IS}
\end{equation}

\begin{equation}
    \rho(\tau) = \Pi_{t=0}^{T}\rho(a_t | s_t)  = \Pi_{t=0}^{T}\frac{\pi^e(a_t | s_t)}{\pi^b(a_t | s_t)} 
    \label{eq:imp_weights}
\end{equation}

\begin{equation}
    G(\tau) = \sum_{t=0}^{T} \gamma^t r_t
    \label{eq:return}
\end{equation}

An important weakness of the IS estimator is its high variance. Trajectories which have a high chance of occurring under the evaluation policy but are rarely observed under the behavioural policy have a high importance weight. Similarly, trajectories which have a low chance of occurring under the target policy contribute little to the value estimate. As a result, the effective sample size used to construct the value estimate is much smaller than the total number of samples used for evaluation. The IS estimator is thus highly sensitive to differences in training samples. Variants of the importance sampling estimator trade unbiasedness for lower variance. To decrease the effect of pathologically large or small importance weights, the weighted IS estimator normalizes the importance weights such that $\sum_{a\ \in A} \rho(a \,|\, s) = 1$. The WIS estimator is not unbiased, but it distributes the weights more smoothly among samples from the training set. Komorowski {\em et al.} ~\cite{komorowski2018artificial} used a WIS estimator to estimate the values of learned policies for the management of sepsis. Following the guidelines of Gottesman {\em et al.} ~\cite{gottesman2018evaluating}, the authors learned a policy for each of the 500 different clustering solutions of the state space, and chose the policy that maximized the 95\% lower confidence bound of the policy's estimated value. 

Direct  methods (DMs) learn the value of the target policy,  either by learning the state transition probabilities or by modeling the value $V^{\pi^e}$ or action-value $Q^{\pi^e}$ function of the evaluation policy. Direct methods suffer from bias due to the difference in state visitation probabilities under different policies, but are more consistent than IS estimators \cite{mrdr}. Doubly robust (DR) methods combine importance sampling and direct method estimates to obtain an estimate with lower variance \cite{jiang2015doubly, mrdr}. 

\begin{equation}
    V_{DR}^{\pi_e}(s_t) = \hat{V}(s_t) + \rho_t(r_t + \gamma V_{DR}(s_{t+1}) - \hat{Q}(s_t, a_t))
    \label{eq:dr}
\end{equation}

The DR estimator, shown in Equation \ref{eq:dr} can be understood as applying a form of bias correction to the DM estimate $\hat{V}(s_t)$, where the bias is estimated using the term $\rho_t(r_t + \gamma V_{DR}(s_{t+1}) -  \hat{Q}(s_t, a_t))$. The importance weight $\rho_t$ is used in this context since the transitions used to estimate the bias are generated from the behavioural policy. Generally the behavioural policy must be estimated from the recorded experiences, the error in its estimate introduces bias in the DR estimator. Since the DR estimator uses an IS estimator, its performance also depends on the effective sample size of the data after applying importance weights. One approach to obtaining value estimates with higher confidence is to constrain the learned policy to have common support with the observed policies. A similar approach was used by Fujimoto {\em et al.} ~\cite{fujimoto2018} to avoid extrapolation errors in batch reinforcement learning, where the agent was constrained to choosing from a set of previously taken actions at a given state. 

\section{Reward design in healthcare}\label{sec:reward_design}

As mentioned in Section \ref{sec:rl}, a reinforcement learning agent is trained to maximize the sum of discounted rewards. The reward function, specified in terms of the current state or state-action pair, should reflect how favourable the state transition is with respect to the underlying goal. For example, a bipedal robot learning to walk should be rewarded for standing upright while displacing in the forward direction. Ideally, by learning to maximize the total reward for a sequence of state transitions, the robot learns a walking gait. Alternatively, the reward could be defined in terms of forward displacement, in which case the robot may learn to achieve a high reward using the more stable means of crawling. Amodei {\em et al.} \cite{DBLP:journals/corr/AmodeiOSCSM16} describe unintended phenomena stemming from poor reward design. Negative side effects are the result of reward functions that do not fully capture our objectives or the restrictions on allowed behaviour. A form of mis-specified reward is using a metric for success that is correlated with successful actions but does not necessarily imply that a successful action was taken.  For example, in \cite{boatrace}, an agent was rewarded for collecting points along a racetrack while the goal of the game was to win the race. Targets could be respawned at one location along the racetrack, so the agent learned to move in a circle, hitting the targets as they respawned, collecting more points while never finishing the race. Reward hacking refers to unexpected policies learned from mis-specified reward functions. Specifically, reward hacking occurs when an agent finds a way to exploit its reward channel to obtain a much higher reward signal than the true reward signal. Since the agent's actions allow it to control what it observes, it may choose to blind itself to observations that provide a low reward.  Current work on RL in healthcare has restricted the agent's action space to specific aspects of treatment, such as managing sepsis through fluids and vasopressors \cite{komorowski2018artificial, raghu2017continuous, raghu2017deep, raghu2018}, or determining policies for weaning a patient from mechanical ventilation \cite{prasad2017reinforcement}. A more diverse action space adds challenges of its own. For example, an agent that is additionally tasked with determining which tests to perform to track a patient's state might choose to avoid receiving negative test results, in order to avoid negative reward. For example, an agent may avoid suggesting regular blood tests to avoid receiving evidence of a worsening infection. This is in contrast with negative side effects generally, in which the designed reward function does not fully express what the designer intended. 

When a health practitioner makes decisions regarding the treatment of a patient, the patient usually trusts that the practitioner is motivated to improve the patient's health. Progress is usually tracked by clinical measures, such as vital signs or test results. When choosing a reward function, a designer must translate high-level objectives to a function of the current state and action. The designer understands objectives expressed in natural language or in demonstrations, but cannot always determine whether an agent trained with a given reward signal will achieve the desired goal. While domain knowledge is an integral part of reward design, it does not rule out the possibility of mis-specified rewards which can result in undesired behaviour. The learned behaviour of an agent is a result of training experience, the environment's representation and dynamics, as well as the learning algorithm and reward function.  While the reward function might be inspired by how we track progress on a given task, its combination with the underlying dynamics and missing experience might lead to unexpected behaviour. Reward design is a well-known challenge in the reinforcement learning community, which needs to be addressed before reinforcement learning can be applied in safety-critical settings such as healthcare. 

\subsection{Reward shaping in healthcare}
 
 In healthcare, some goals can be expressed as binary reward functions, such as preventing patient mortality from sepsis \cite{komorowski2018artificial}, where the reward is given at the end of a patient's trajectory, i.e.,  $+100$ for survival and $-100$ for mortality. Binary rewards, however, preclude the expression of secondary objectives such as minimizing length of stay and total treatment costs. Additionally, binary rewards increase the sample complexity of a reinforcement learning problem, and aggravate difficulties such as the credit assignment problem. 
It is possible to transform binary or sparse rewards, such that the optimal policy under the original reward function is the same as the optimal policy under the modified reward function. Ng {\em et al.} \cite{ng1999policy} derive a general form for intermediate rewards for which the optimal policy is the same as that of the corresponding binary or sparse reward function. The general form, termed potential-based reward shaping, is described in Equations \ref{eq:potential} and \ref{eq:reward_shaping}, where $\phi(s)$ can be any chosen function of the state. Equation \ref{eq:potential} describes the form of the potential function $F(s, a, s^{\prime})$ required to ensure that the reward transformation is policy-invariant. Without further knowledge of the transition dynamics, Equation \ref{eq:reward_shaping} is the only type of transformation that can guarantee policy invariance. 
 
\begin{equation}
F\left(s, a, s^{\prime}\right) = \gamma \Phi\left(s^{\prime}\right)-\Phi(s) 
\label{eq:potential}
\end{equation}

\begin{equation}
    R^{\prime}(s, a,  s^{\prime}) = R(s, a, s^{\prime}) + F(s, a, s^{\prime}) 
    \label{eq:reward_shaping}
\end{equation}
While Equations \ref{eq:potential} and \ref{eq:reward_shaping} result in a policy-invariant reward transformation for any function $\phi(s)$, domain knowledge is generally needed to obtain the benefits of lower sample complexity \cite{ng1999policy}. For instance, $\phi(s)$ could be an estimate of the risk of mortality given the patient's current state. Raghu {\em et al.} \cite{raghu2017continuous,raghu2017deep} use intermediate rewards for sepsis management, based on clinical measures of severity, while the intended measure of success is mortality. Peng {\em et al.} \cite{peng2018improving} use a mortality predictor to obtain intermediate rewards. However, both reward formulations do not adhere to the potential-based reward transformation shown in Equation \ref{eq:potential}. The effects of such miss-specified intermediate rewards are discussed in Section \ref{sec:sepsis}.  

\subsection{Multi-Objective Rewards}
While most applications of reinforcement learning in healthcare focus on treatment decisions, Cheng {\em et al.} \cite{cheng2019optimal} use reinforcement learning to obtain a policy for ordering phlebotomy tests for patients suffering from sepsis or acute renal failure. Since phlebotomy tests are often costly and can result in effects such as low hemoglobin levels \cite{loftsgard2016clinicians}, a policy for ordering tests should balance the need to track the patient's state while minimizing healthcare costs and patient discomfort. Preferences for such a trade-off are not easily expressed with a single reward function. The authors use fitted Q-iteration (FQI) with a multi-objective or vector-valued reward as shown in Equation \ref{eq:vector_reward}, where $r^{SOFA}$ penalizes increases in SOFA scores, $r^{treat}$ denotes a positive reward if a treatment was started after a lab test was ordered, $r^{info}$ rewards test results that are informative and $r^{cost}$ denotes the monetary cost of the lab test. 

\begin{equation}
\boldsymbol{r}_{t}=\left[r_{t}^{S O F A}, r_{t}^{t r e a t}, r_{t}^{i n f o},-r_{t}^{c o s t}\right]^{\top}
\label{eq:vector_reward}
\end{equation}
To estimate the information gain needed for $r_t^{info}$ the authors train a multi-output Gaussian process to predict hourly values of future measurements such as vital signs and test results. $r^{info}$ is computed using the absolute difference between predicted measurements according to the Gaussian process regression model, and measurements received from the lab results. 

In the case of multiple objectives or vector-valued rewards, the notion of optimal policies is replaced by the set of Pareto-optimal or un-dominated policies. A Pareto-optimal policy is a policy which cannot be changed without incurring a lower return in at least one objective. 
Pareto-optimal policies do not necessarily incur a high return for all objectives, especially in the case of conflicting objectives where policies must account for a preferred trade-off. One can define a scalarization function to transform the vector of rewards to a scalar, according to preferences or relative importance of each objective. For example, scalar rewards can be formed from convex combinations of the objectives. The optimal policy for a given scalarization function is a Pareto-optimal policy, however not all Pareto-optimal policies correspond to optimal policies for some convex combination of reward components. Cheng {\em et al.} \cite{cheng2019optimal}, use a variation of Pareto-optimal fitted Q-iteration (PO-FQI) \cite{lizotte2016multi} to find an optimal policy that conforms to a clinician's preference. The method introduced by Lizotte {\em et al.} \cite{lizotte2016multi} finds a set of un-dominated actions $\Pi(s)=\left\{a : \nexists a^{\prime}\left(\forall d, \hat{Q}_{d}(s, a)<\hat{Q}_{d}\left(s, a^{\prime}\right)\right)\right\}$ for each state in the state space at each time step of the horizon. The map $\Pi: S \rightarrow A$ describes a \textit{non-deterministic policy} (NDP) \footnote{The term non-deterministic does not mean a stochastic policy in this context} since each state is mapped to a set of actions instead of a single action or a single distribution over actions. A NDP thus represents a set of possible policies. The algorithm for PO-FQI proceeds similarly to FQI but learns a set of approximations of a vector-valued Q-function, where each approximation corresponds to a policy that is consistent with the NDP $\Pi$. 

In order to ensure that PO-FQI is tractable, in addition to consistency with $\Pi$, the condition of representability is imposed to further limit the set of policies to be considered. For instance if the Q-function approximator is a linear function with respect to state-action features $\phi(s, a)$, then $\pi_t(s_t) = argmax_a \phi(s_t, a)^T \textbf{w}$ for some real vector $\textbf{w}$. This requires a previously obtained state-action representation $\phi(s, a)$, which can be hand-crafted or learned. Moreover, in the case of a binary action space, such as whether to order a lab test, both actions might be Pareto-optimal and representable, which means that the set of policies learned by PO-FQI is not getting pruned. Cheng {\em et al.} \cite{cheng2019optimal} use a stricter pruning scheme for un-dominated actions shown in Equation \ref{eq:pruning}. A policy is learned for each lab separately, where the action space is binary, with $a = 1$ if the lab test should be ordered. Specifically, a lab test is ordered if the expected return, in addition to a preference variable $\epsilon_d$, is greater than the null action along all dimensions of the reward vector. 

\begin{equation}
\Pi(s)=\left\{\begin{array}{ll}{1, \hat{Q}_{d}(s, a=0)<\widehat{Q}_{d}(s, a=1)+\varepsilon_{d},} & {\forall d} \\ {0, \text { otherwise }}\end{array}\right.
\label{eq:pruning}
\end{equation}
 The hyperparameters $\epsilon_d$ describe the relative priority of each objective. For instance if cost is not a primary concern then $\epsilon_{cost}$ should be set to a positive value. For more critical objectives, $\epsilon_d$ could be set to a negative value.  Cheng {\em et al.}\cite{cheng2019optimal} tune $\epsilon_d$ so that the total number of orders for each lab test under the new policy is close to the number of tests ordered by clinicians in the training set. To comply with clinical protocols, the authors introduce a budget that suggests taking a lab test if the test has not been ordered in the last 24 hours.
 
While multi-objective rewards can better represent the clinicians' intents and preferences, they are still amenable  to misspecification. For instance, Cheng {\em et al} include the SOFA score in the set of objectives, however the action space only consists of lab orders, which do not directly affect the SOFA score. Since treatment decisions, which influence the SOFA score, are not incorporated into the action space nor the state space, they are considered unobserved confounders. For instance, patients with a high SOFA score might be monitored more frequently, thus clinicians might order more lab tests and the SOFA score might remain high. Such trajectories in the training set wrongly penalize frequent testing for severely ill patients if the SOFA score is considered an outcome or a penalty for lab test orders. 

The authors evaluate the learned policy using per-step weighted importance sampling (PS-WIS) and compare its estimated value with that of the behavioural (clinician's) policy and three random policies which order tests with probabilities 0.01, $p_{emp}$ and 0.5, where $p_{emp}$ was the empirical probability of ordering a lab test.  PS-WIS often estimates higher values for the random policies than that of the behavioural policies. For instance, the random policies, including that which orders tests with probability 0.01, are estimated to result in a greater expected information gain than the behavioural policy. Additionally, they are estimated to result in a change of treatment more frequently than the behavioural policy. These results are conspicuous since the behavioural policy orders tests more frequently than the random policy with probability 0.01, and should therefore obtain greater information gain. Like any WIS evaluation method, PS-WIS requires an estimate of the behavioural or clinician's policy which the authors learned using the training data. It is worth noting that lab orders are sparse, and for most observed states, no lab was ordered. This dataset imbalance could lead to a poor estimate of the clinician's policy, predicting a low probability of ordering a test at every state, which could then result in misleading value estimates given by the PS-WIS method. It is therefore imperative that models of the behavioural policy are properly validated before using them in OPE methods. 

\subsection{Learning objectives from demonstrations}

An alternative to handcrafting reward functions is to learn them from demonstrations. Inverse reinforcement learning (IRL) enables an agent to learn a reward function from sample trajectories, assuming that the sample trajectories demonstrate the behaviour of an optimal policy with respect to the hidden reward function. IRL is especially useful when it is difficult to express the reward function in terms of the many relevant features. For example, Yu {\em et al.} \cite{yu2019inverse} use Bayesian IRL to learn a reward function for mechanical ventilation weaning and optimal sedative dosing in the ICU. The action space consists of a binary action indicating whether a patient should remain on mechanical ventilation, along with four levels of sedative dosage, for a total of 8 possible action combinations. The reward function is expressed as the sum of three reward criteria $r = r^{vitals} + r^{vent on} + r^{ventoff}$. The first term introduces a penalty if vital signs are outside the normal range or if vital signs undergo sudden changes. The second term represents the cost of each additional hour spent on the ventilator, and the third term rewards successful extubations while penalizing unsuccessful extubations. The reward function is parameterized by seven parameters $[C_1, \ldots, C_7]$ which control the trade-off between individual criteria of the reward. 

Learning the reward function for which the observed behaviour is optimal is an ill-posed problem, since the observed policy may be optimal for many reward functions. However, one can compute a probability distribution over the space of possible reward parameters. Let $P(O \,|\, R ; P)$ denote the probability that a sequence of state-action pairs $O$ is generated by an agent acting optimally in an environment with transition probabilities $T$, with respect to a reward function parameterized by $R$. Using Bayes' rule, the probability distribution over reward parameters can be written in terms of the likelihood of observed sequences $P(R \,|\, O ; T) = \frac{P(O \,|\, R ; R)P(R)}{P(O)}$. An optimal policy, according to the recovered reward function, must induce the same expected return as the expert's policy. This fact provides a first moment constraint on the likelihood function. The maximum entropy distribution with a first moment constraint is the Boltzmann distribution shown in Equation \ref{eq:energy}. For interpretability, the function $f(O, R)$ can be expressed as a linear combination of features of the patient's trajectory $O$, such as total length of stay, vital signs and risk scores. The parameters of the reward function $R$ can be learned by maximizing the likelihood $P(O \,|\, R ; T)$ \cite{ziebart2008maximum} or a MAP estimate $P(O \,|\, R ; T)P(R)$ \cite{ramachandran2007bayesian}.  

\begin{equation}
    P(O | R ; T) = \frac{e^{f(O; R)}}{Z(R)}
    \label{eq:energy}
\end{equation}

In Bayesian IRL \cite{ramachandran2007bayesian}, the likelihood of an expert trajectory $Pr(O | R)$ for a given reward function $R$  is given by the Boltzmann distribution $e^{\alpha \sum_{i} Q^{*}\left(s_{i}, a_{i}, R\right)}$. The probability of the reward function parameters is obtained from a MAP estimate shown in Equation \ref{eq:map}
\begin{align}
\label{eq:posterior}
\operatorname{P}(O | R) & =\frac{1}{2} e^{\alpha \sum_{i} Q^{*}\left(s_{i}, a_{i}, R\right)}\\
\label{eq:map}
R^* & = argmax_R \operatorname{P}(O | R)\operatorname{P}(R)
\end{align}

where the observed trajectory $O$ consists of state-action pairs $(s_i, a_i)$, the prior $Pr(R)$ is a uniform distribution over the interval $[0, 1]$ for each constant in $[C_1, ... C_7]$  and $Q^*(., ., R)$ is the action-value function for the optimal policy under reward parameters $R$.  A form of MCMC algorithm termed PolicyWalk \cite{ramachandran2007bayesian} is used to estimate the mean of the posterior $Pr(R | O)$ which proceeds as follows: at each iteration, a new hypothesis reward $\tilde{R}$ is uniformly sampled from the neighbors of the current hypothesis $R$. Yu {\em et al.} \cite{yu2019inverse} use FQI to compute the Q-function and the optimal policy with respect to reward parameters $R$ which is then used to obtain the posterior ${\operatorname{Pr}(\tilde{\mathbf{R}} | O)}$ in accordance with Equation \ref{eq:posterior}. The newly sampled reward $\tilde{R}$ is accepted with probability $\min \left\{1, \frac{\operatorname{Pr}(\tilde{\mathbf{R}} | O)}{\operatorname{Pr}(\mathbf{R} | O)}\right\}$.  Thus the Bayesian IRL algorithm returns the parameters of the inferred reward as well as its optimal policy. 

% Evaluation 
Yu {\em et al} \cite{yu2019inverse} compare the performance of the policy learned using Bayesian IRL $\pi_{BIRL}$ with several baseline policies. The first baseline policy $\pi_{BL}$ was learned using FQI with $[C_1, ... C_7]$ set to $[1/7, 1/7, 1/7, 1/7, 1/7, 1/7, 1/7].$ The remaining baseline policies $\pi_{BL_1} \ldots \pi_{BL_3}$, $[C_1, \ldots, C_7]$ were set to uniformly sampled non-negative values that sum to 1. Both $\pi_{BIRL}$ and  $\pi_{BL}$ were consistent with clinicians' ventilation actions 99.6\% and 99.7\% of the time, respectively. This is unsurprising since ventilation actions in the training set are binary and sparse, thus precision and recall are required to properly evaluate the performance of the learned reward parameters and policies. The remaining baseline policies  $\pi_{BL_1},  \pi_{BL_2}, \pi_{BL_3}$ achieve much lower rates of consistency since the constants pertaining to the reward terms $r^{ventoff}$ and $r^{vent on}$ are randomly set to high or low values in the $[0, 1]$ interval which may disproportionately penalize ventilation costs or unsuccessful extubations. Both $\pi_{BIRL}$ and  $\pi_{BL}$ also achieve similar consistencies with clinicians' sedative dosing actions, with respective matching rates of  54.2\% and 53.5\%, while the remaining baseline policies $\pi_{BL_1},  \pi_{BL_2}, \pi_{BL_3}$ achieve much lower consistency rates. Despite achieving similar consistency rates with $\pi_{BL}$, the reward parameters for which $\pi_{BIRL}$ is optimized are very different from the reward parameters of $\pi_{BL}$, suggesting that similar policies may be obtained from a large range of reward parameters. In this case, it may be due to correlations between the objectives of the reward function. For instance, the stability of vital signs is disrupted by unsuccessful extubations, which in turn might result in more hours on the ventilator. We reiterate that each iteration of IRL requires learning the Q-function of the optimal policy with respect to the current estimate of the reward parameters, this makes IRL susceptible to extrapolation errors which are commonly encountered in off-policy RL from observational data. To alleviate this problem, the PolicyWalk algorithm may be modified by imposing a restriction on the policies of accepted reward hypotheses $\tilde{R}$. For each policy  returned by the FQI step,  the corresponding reward hypothesis $\tilde{R}$ is rejected if at certain states the policy samples actions which are never or rarely taken by clinicians, regardless of the estimated Q-function. Moreover, if the environment dynamics are unknown, IRL suffers from an inherent ambiguity. If clinicians are treating a difficult case, the patient's trajectory may reflect low returns due to unfavourable outcomes and higher costs. Such trajectories when used as expert demonstrations may unduly influence the estimate of the reward function. 

Learning a reward function by treating medical records as demonstrations of optimal behaviour undermines the goal of learning better policies than current practices. 
A naive approach is to select from the training set only those trajectories which result in a favourable outcome, thus eliminating sample demonstrations which we may not want the RL agent to emulate. However this severely limits the set of states present in the training set, and may restrict the set of sample demonstrations to less severe cases of illness. Since the RL agent's policy must not deviate too much from the clinician's policy, the reward function parameters for which an improved policy is optimal are expected to be close to the parameters for which the clinicians' policy is optimal. Hence IRL may still provide a useful starting point for reward design. 

One could also characterize different approaches by their respective reward functions. If observed records are generated by different physicians or hospitals, the records may reflect different approaches or decision-making preferences. For example, a clinician who prescribes treatments earlier might prefer more preventative strategies than clinicians who choose to observe patients before taking action, and clinicians who prescribe more diagnostic tests might be more risk averse at the expense of higher costs. A learned reward function might reflect such hidden preferences. Equation \ref{eq:energy} suggests that one can consider the parameters of a reward function as a latent variable that explains the state-action pairs generated by an agent. Different approaches could be represented by different parts of the latent space. The learned reward functions could be examined to obtain a reward function that better captures our notion of desired behaviour. 

\section{State Representation and Model-Based RL}\label{sec:StateRep}

Reinforcement learning algorithms assume a Markov model of the agent's environment. This assumption is not restrictive if the choice of state representation is such that the next state can be predicted only from the current state and action. Essentially, the chosen state representation must be a relevant summary of the patient's history. Recurrent neural networks can learn such a representation by being trained to predict the outcome or the next observation given a sequence of observations and actions. The learned state representation can be considered as a \textit{patient embedding}, where patients of a similar phenotype (or having similar predicted outcomes) obtain similar state representations. This idea was employed by Zhang {\em et al.}~to learn a vector representation of a patient's EMR that was then used to predict the risk of hospitalization \cite{zhang2018patient2vec}. 

The predictive model that learns a state representation can also be used to learn an optimal policy. Model-based reinforcement learning makes use of a predictive model of the environment to choose actions over several future time steps, and adjusts the plan if the outcome does not meet the predictions. However, the model can suffer from a distributional shift in state transitions when the agent is generating experiences under a different policy. The agent can exploit vulnerabilities in the predictive model to come up with optimistic plans. For example, Oberst and Sontag \cite{oberst2019} showed that transition probabilities estimated from observational data with hidden confounders led to wrong predictions of discharge under the RL policy. Their experiments are discussed in Section \ref{sec:causality}. 

 Modeling the distribution of the next state instead of predicting point estimates is preferred since the uncertainty of the predictive model can be used to make the RL agent risk-averse as was shown by Depeweg {\em et al.}\cite{depeweg2017uncertainty} on the Siemens industrial benchmark \cite{hein2016introduction}. The authors trained a Bayesian neural network to predict state transitions, and learned an approximate posterior $q(W)$ of the weights of the network, given the training data. The authors then extended the reward function with a term quantifying the uncertainty of the expected reward as a result of the uncertainty over the weights $W$, and showed that a policy optimizing the extended reward is more predictive of performance at test time.  A similar approach might help RL algorithms avoid learning degenerate policies due to missing experiences in the training data, such as a lower proportion of severe cases or a limited range of observed actions.

\section{Application to Sepsis Management}\label{sec:sepsis}
Several papers propose to use reinforcement learning to optimize sepsis management in the ICU. Training data is obtained from MIMIC-III \cite{johnson2016mimic}. Treatment for sepsis may involve antibiotics, intravenous fluids, vasopressors, and mechanical ventilation \cite{sepsis_guide}. However, current work restricts the action space to the set of combinations of IV and vasopressor dosages, since a large action space requires a larger training set to learn and evaluate a new policy. Observed IV and vasopressor dosages are discretized into 5 bins each, for a total of 25 distinct actions. 

% Different objective, in-hospital vs 90d mortality
Komorowski {\em et al.}~\cite{komorowski2018artificial} defined the reward function in terms of 90-day mortality. A trajectory that ended with mortality was given a reward of $-100$, with a discount factor of 0.99. A high discount factor penalizes late deaths almost as much as early deaths. A trajectory that ended with survival after 90 days of the onset of sepsis was given a reward of 100. Raghu {\em et al.}~\cite{raghu2017continuous} also use a positive reward ($+15$) for survival and a penalty ($-15$) for death at the end of a patient's trajectory. Compared to other work \cite{raghu2017deep,raghu2017continuous} where the reward was based on the SOFA score, binary rewards are simpler and less prone to misspecification, however their sparsity decreases the sample efficiency of reinforcement learning algorithms and increases the variance of off-policy value estimates. 

Raghu {\em et al.}~\cite{raghu2017continuous} \cite{raghu2017deep}, define intermediate rewards based on the SOFA score, changes to the SOFA score and lactate levels, as shown in Equation \ref{eq:sofa_reward} with a reward of $+15$ or $-15$ at the end of a patient's trajectory.

\begin{equation}
r\left(s_{t}, s_{t+1}\right)=C_{0} \mathbb{1}\left(s_{t+1}^{\mathrm{SOFA}}=s_{t}^{\mathrm{SOFA}} \& s_{t+1}^{\mathrm{SOFA}}>0\right)+C_{1}\left(s_{t+1}^{\mathrm{SOFA}}-s_{t}^{\mathrm{SOFA}}\right) +
C_{2} \tanh \left(s_{t+1}^{\mathrm{Lactate}}-s_{t}^{\mathrm{Lactate}}\right)
\label{eq:sofa_reward}
\end{equation}

While the reward function was clinically motivated, Raghu {\em et al.}\cite{raghu2017deep} observed that the learned policy recommended lower dosages of vasopressors and IV fluids for severely ill patients.  As mentioned by Gottesman  {\em et al.} \cite{gottesman2018evaluating} this might be due to fitting a Q-function with missing state-action pairs. In a later work, Raghu {\em et al.}~\cite{raghu2018} used a similar reward function but ensured that the learned policy did not differ too much from the clinician's policy by setting a small learning rate for the policy gradient. 

Peng {\em et al.} \cite{peng2018improving} design an intermediate reward based on the likelihood of patient mortality at the end of the trajectory. The authors train a regression model that predicts the probability of mortality given a patient's current observations. The reward function was defined as shown in Equation \ref{eq:peng_reward} where $f(o)$  is the probability of mortality, as predicted by the regression model, given current observations $o$. The reward function penalizes increases in the log-likelihood of mortality given the observations of the next time-step $o'$.  One problem with such a reward construction is the dependence of the probability of mortality on the behavioural or clinician's policy. Therefore, the likelihood of mortality under the target policy may be different. This might influence the learned policy since the reward function is not consistent with policy-invariant reward transformations as discussed in Section \ref{sec:reward_design}.

\begin{equation}
r\left(o, a, o^{\prime}\right)=-\log \frac{f\left(o^{\prime}\right)}{1-f\left(o^{\prime}\right)} f\left(o^{\prime}\right)+\log \frac{f(o)}{1-f(o)}
\label{eq:peng_reward}
\end{equation}

Komorowski {\em et al.}~\cite{komorowski2018artificial} extracted vital signs, laboratory values, clinical scores and demographics of each patient. To reduce the size of the state space, the authors clustered the patient's measurements using k-means++ to obtain 750 discrete states. A larger number of clusters provides higher granularity but many states would be sparsely observed while training. Komorowski {\em et al.}~\cite{komorowski2018artificial} verify the validity of the learned state representation by examining the distribution of ICD codes in the state clusters to determine whether past medical history and diagnoses were included in the state representation.  In contrast, Raghu {\em et al.}~\cite{raghu2017continuous} learned a lower-dimensional continuous state representation using a sparse autoencoder, and learned a policy using deep Q-learning. When using function approximators, it is important to ensure that patients with similar trajectories or demographics are assigned similar states, as the Q-function is expected to interpolate between the observed trajectories.  Additionally, the Markov property must be verified for the learned state representation since it must be sufficient for predicting the next state.  Komorowski {\em et al.}~\cite{komorowski2018artificial} used 80\% of the development set to estimate the transition probabilites using normalized transition frequencies. To verify the Markov property, the authors simulated 500 random walks using the learned transition matrix, and counted the number of times an agent would remain in a given state for a given number of time-steps. If the state transition process is memoryless, the life-expectancy for each state must be exponentially distributed. The authors verify the correlation coefficient between the data and an exponential decay function fitted on the simulated data. Policy iteration was used to learn a new policy using the estimated transition probabilites. The learned policy recommended a higher dosage of vasopressors than clinicians. Patients from the test set would have received vasopressors 30\% of the time under the learned policy, but received vasopressors 17\% of the time under the clinicians' policy. Specifically, higher vasopressor dosages were recommended for patients who also received higher IV fluid dosages. The authors point to clinical literature which suggests that early administration of vasopressors might improve outcomes \cite{byrne2017fluid, marik2015demise}. Raghu {\em et al.}~\cite{raghu2017continuous} also found that the learned policy recommended vasopressors more often than clinicians, but mostly in cases where the administered IV fluid dosage was low. Careful examination of the trajectories that recommended a change of dosage is required in order to understand whether the recommended changes are due to extrapolation errors. Gottesman {\em et al.} describe a possible side effect of limiting the action space. When an agent was restricted to only consider when to intubate a septic patient, the learned policy recommended intubation more frequently than physicians. This might be a case of extrapolation errors common with off-policy batch RL, since the agent cannot recommend changes to other treatment strategies, and may not have observed the negative effects of an aggressive intubation policy. 

Peng {\em et al} \cite{peng2018improving} use an LSTM autoencoder to obtain a 128-dimensional state representation based on patient history. The autoencoder is trained to minimize the reconstruction error between the original measurements and the decoded measurements. However, the authors do not examine the state representation for correlations with measurement time, clinical severity scores or demographics. The autoencoder was used to represent the state of all patients in the training set, at 4-hour intervals. For each state vector $s_t$, 300 Euclidean nearest neighbors were found. The kernel policy $\pi_k$ is the distribution of actions taken by clinicians among the survivors of the nearest neighbors. The authors found that the kernel policy exhibited a bias towards recommending no action, and postulate that this might be due to the fact that surviving neighbors might have been healthier and required fewer interventions. The state representation is meant to cluster similar patients, with similar measures of severity more closely so that actions for a more severely ill patient are not influenced by those taken for healthier patients. The bias of the kernel policy towards recommending no action suggests that the Euclidean distance between learned state representations does not reflect patient similarities. Unlike the kernel policy, the DQN model exhibited a bias towards more aggressive interventions. The authors use the kernel policy $\pi_k$ in a mixture-of-experts model, which selects between the action recommended by $\pi_k$ and that of a Deep Q-Network (DQN) model in order to maximize the expected return as estimated by a weighted doubly robust estimator. 

While the above proposed solutions to optimizing sepsis management report greater estimated returns and lower expected rates of mortality, the authors also acknowledge the limitations of such approaches. Since design choices must be made when modeling the state space, action space, specifying the reward function and choosing the RL algorithm, several factors may influence the learned policy. Off-policy evaluation methods are also influenced by the choices of state and action representations and the specified reward function, and must therefore be supplemented by examinations of the state representation, and restrictions on the set of actions that an agent may recommend. Additionally, it is imperative to verify the effective sample complexity when using WIS or DR evaluation methods as discussed in Section \ref{sec:rl_obs}. 

\section{Application to Heparin Dosing}\label{sec:heparin}

Nemati {\em et al.} \cite{optimalheparin} and Lin {\em et al.} \cite{lin2018deep} use deep reinforcement learning to learn an optimal policy for intravenous heparin dosing. Heparin is an intravenous (IV) anticoagulant used to treat blood clots among patients with severe renal failure because other anticoagulants are contraindicated. One difficulty is that of sparse measurements and sparse rewards. Since a blood test is required to measure aPTT, observations are not frequently updated, and the response of actions is delayed. 

Nemati {\em et al.} \cite{optimalheparin} use a discriminative hidden Markov model (DHMM) to represent the patient's state conditioned on a history of observations and actions. The output of the DHMM as well as the current action taken are passed to a Q-network. Both the DHMM's and the Q-network's parameters are updated by minimizing the TD-error.  Thus, a state representation is learned from patient history alongside the Q-function. While such an approach provides an end-to-end solution, it precludes the verification of the learned state representation before learning an improved policy. Additionally, under this training scheme, the learned state representation is only used to predict the action value and may not be as informative as state representations which are learned via clustering, autoencoding or next-state prediction methods. 

Lin {\em et al.} \cite{lin2018deep} state that unlike sepsis management with IV fluids and vasopressors, heparin dosing policies cannot be learned with a discrete action space, since precise dosage control is required to maintain the activated partial thromboplastin time (aPTT) within the therapeutic range for patients receiving heparin. The authors use the deep deterministic policy gradient (DDPG) algorithm \cite{ddpg} to learn an optimal policy over continuous state and action spaces. The deterministic policy is parameterized by a neural network that outputs the recommended heparin dose given patient measurements at a given time. The DDPG algorithm is an instance of actor-critic algorithms discussed in Section \ref{sec:rl}. The critic is parameterized by a neural network which estimates the action-value or Q-function of the policy network. The training and test sets consist of electronic patient data from the MIMIC-III dataset \cite{johnson2016mimic} and the Emory hospital intensive care unit. Both Nemati {\em et al.} ~\cite{optimalheparin} and Lin {\em et al.} \cite{lin2018deep} used the reward function defined in Equation \ref{eq:heparin_reward} for the objective of maximizing the fraction of time a patient's aPTT stays within the therapeutic range of heparin, which is between 60 and 100 seconds according to the Beth Israel Deaconess Medical Center \cite{ghassemi2014data}. The reward value is 1 for aPTT values within the therapeutic range, and quickly diminishes to $-1$ as the aPTT values move further from the therapeutic range. 

\begin{equation}
   r_t = \frac{2}{1 + e^{-aPTT_t + 60}} - \frac{2}{1 + e^{-aPTT_t + 100}} - 1 
   \label{eq:heparin_reward}
\end{equation}
Lin {\em et al.} \cite{lin2018deep} use a direct method to evaluate the learned policy. The authors define a distance between the RL policy and the clinicians' policy over a patient's trajectory as the mean difference between the recommended dose and the administered dose. The policy distances are binned into 5 quantiles. For each quantile, the authors plot the observed return of the clinicians' policy, showing decreasing returns as the policy distance increases and that the expected return is greatest when the clinicians' policy matches the RL policy. Since high return trajectories are more likely to occur in less severe cases, it is difficult to ascertain whether the learned policy is better than the clinicians' policy. The authors perform a multiple linear regression analysis using patient characteristics, clinical severity scores and policy distance as features, and the expected return as the outcome.  The magnitude of coefficients of patient characteristics and severity scores (SOFA and pulmonary embolism) were close to 0 and had high $p$-values while the coefficient for policy distance had a greater magnitude and a low $p$-value ($p < 0.05$). While such an analysis appears to show that the expected return is mainly determined by the policy distance and is independent of severity, statistical tests cannot rule out the possibility of confounding \cite{pearl1998there}, crucially, such an analysis does not validate the recommendations made by the RL policy in areas where the policy distance is large.  Additionally, the policy distances were binned into 5 quantiles, if the distribution of policy distances has a long tail, then effectively most policy distances would correspond to the 0-distance bin.

\section{The Role of Causality} \label{sec:causality}

The construction of better treatment strategies from observational data may be viewed as a problem of causal inference. Reinforcement learning involves choosing actions based on their subsequent effect on the state. In a sense, an agent must learn the effect of its actions in an environment governed by unknown dynamics.  While causal inference seems to be closely tied to reinforcement learning, few papers propose ways to combine their strengths.  In this section we review the basic principles of causal inference and describe its potential role in sequential decision support. A more thorough treatment of causal inference can be found by Peters {\em et al} \cite{peters2017elements}. 

\subsection{Structural Causal Models}
\begin{figure}
\centering
\includegraphics[scale=0.3]{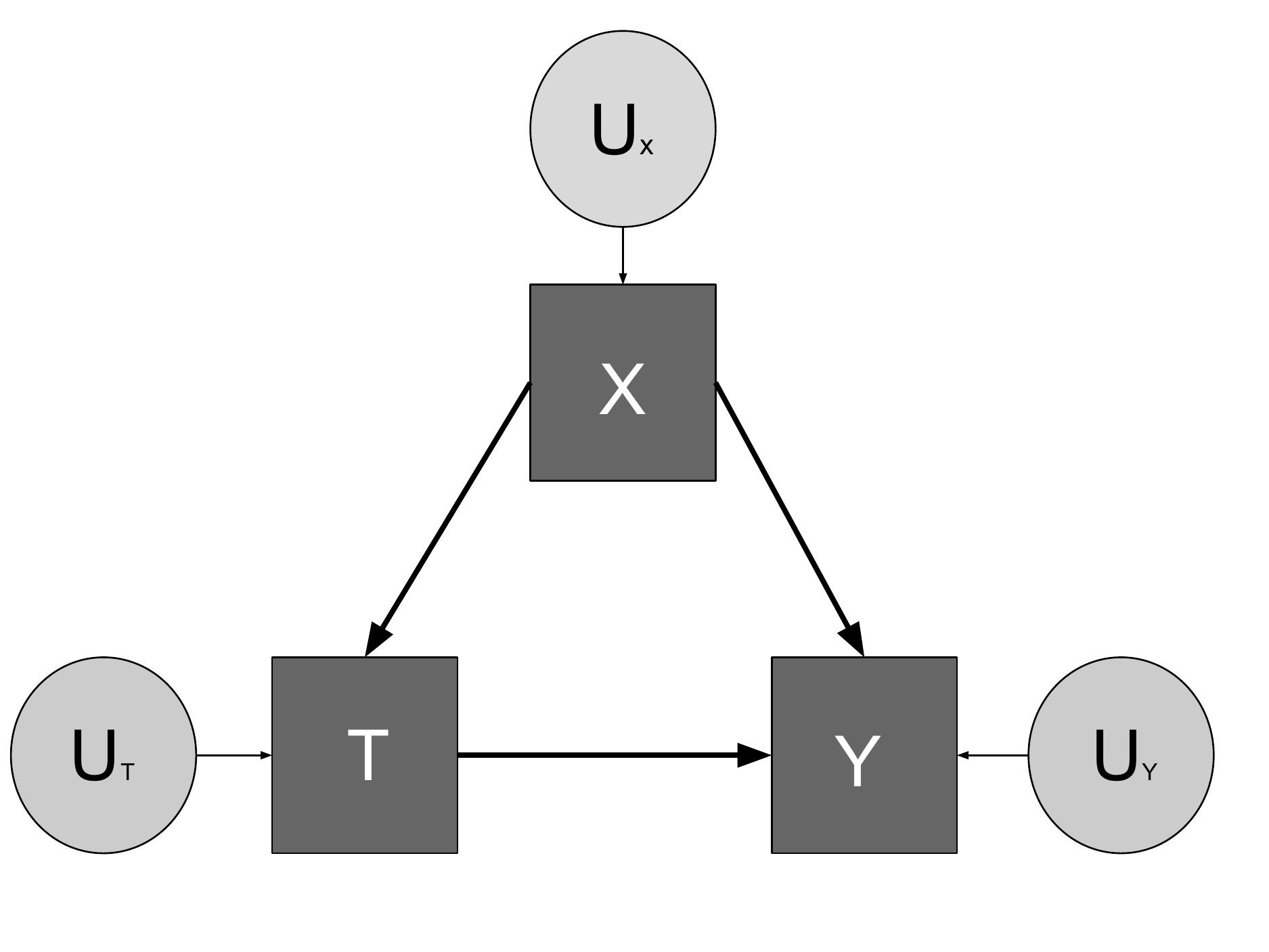}
\caption{A structural causal model of applying treatment $T$, with covariates $X$ and outcome $Y$. Unobservable variables $U_X$,  $U_T$, and $U_Y$ are independent and sampled from a prior distribution $P(\textbf{U})$} 
\label{fig:scm}
\end{figure}

Structural causal models are built on the assumption that  data is generated by an underlying mechanism which entails an observational distribution.  Figure \ref{fig:scm} shows a structural causal diagram or a causal directed acyclic graph (DAG) with three observable variables, the covariates $X$, treatment T, and outcome $Y$.  The latent variables $U_X$, $U_Y$, and $U_T$ are unobserved independent noise variables. The SCM, typically denoted by $M$, describes a data generating process as follows: first the noise variables $U_X$, $U_Y$, and $U_T$ are sampled from their respective prior distributions. Since X is the root node of the observable variables, it is generated first using its structural equation $X := f_x(U_x)$. The treatment $T$ depends on $X$ and its random noise variable $U_T$. The outcome $Y$ depends on the covariates $X$, treatment $T$, and noise variable $U_Y$ through its structural equation $Y := f_y(X, T,  U_Y)$.  Such an SCM entails an observational distribution $P^M(X, Y, T) = P(Y | X, T)P(T | X)P(X)$ which can be factored according to the conditional independencies conveyed by the structural causal diagram. In general, structural causal models describe how variables are generated in terms of their parents in the DAG, that is $X := f_x(\textbf{PA(X)})$ where $\textbf{PA(X)}$ denotes the set of parent nodes of $X$.  Reinforcement learning involves modifying the policy which generated the observed trajectories thus changing the data-generating process. Structural causal models (SCM) provide a framework for describing the data-generating process and the distributions entailed by interventions.

% Need to link back to OPE
An intervention $
I=d o\left(X_{i} :=\tilde{f}\left(\mathbf{P} \mathbf{A}_{i}, \tilde{U}_{i}\right)\right)
$ modifies the SCM by replacing the structural mechanism which generates $X$ by $X :=\tilde{f}\left(\mathbf{P} \mathbf{A}_{i}, \tilde{U}_{i}\right))$  for any observable variable $X$ of the SCM. For instance, changing which features are considered when selecting a treatment $T$ corresponds to changing the set of parent nodes of $T$, and updating the parameters of a policy corresponds to changing the mechanism $f$. An intervention $I$, results in a different SCM $M_I$, and therefore can entail a different distribution $P^{M_I}(X, Y, T)$, termed the interventional distribution. Computing the expected return of a new policy can therefore be seen as computing the expected outcome $Y$ under an intervention. To compute what would have happened to a particular patient under a different policy or particular intervention $I$ requires computing the \textit{counterfactual} distribution. Counterfactual distributions $
P^{\mathcal{M} | \mathbf{X}=\mathbf{x} ; I}$ are evaluated as follows:  the latent variables are first sampled from the posterior distribution $\textbf{U} \sim P(U | \textbf{X} = \textbf{x})$, then the interventional SCM  $M_I$ is used to compute the value of observable random variables, starting with root nodes as described above.

% The non-identifiability problem
Generally, different SCM's can entail the same interventional and observational distributions but different counterfactual distributions. As a result, even if we learn mechanisms that fit the data well, they might be inadequate at predicting counterfactuals. A new policy may have a higher expected return, but it can be impossible to determine for which patients the outcome is improved.  Certain types of SCMs do not have this problem; their counterfactual distributions can be uniquely determined. However, the assumptions required to form such an SCM are untestable. For instance, in the case of binary treatments and outcomes, the monotonicity condition is sufficient to identify counterfactual distributions of a given SCM \cite{pearl2009}. 

Oberst and Sontag  \cite{oberst2019} propose a method of evaluating a policy based on counterfactually generated trajectories. The authors generalize the monotonicity condition to categorical treatments and outcomes, termed the counterfactual stability condition. To sample trajectories with categorical treatments and outcomes, Oberst and Sontag  \cite{oberst2019} use a Gumbel-Max SCM and show that it satisfies counterfactual stability, hence its counterfactual distributions are identifiable. In order to illustrate the use of counterfactual estimation for policy evaluation, the authors build a simplified sepsis simulator where state transitions are modeled using a Gumbel-Max distribution, thus assuming that the true data-generating distribution is counterfactually identifiable. Observations from the simulator consist of four vital signs, heart rate, blood pressure, oxygen concentration and glucose levels, each is discretized into 3 bins; low, medium and high.  The behaviour or clinician's policy is set close to optimal by performing policy iteration using the simulator's MDP and taking a random action with probability 0.05. The authors use 1000 sampled trajectories from the simulator using the behaviour policy as the observational data used for training. The simulator is not used to learn the target policy since the authors wish to illustrate the potential pitfalls of training with observational data. 

\begin{figure}[h!]
\centering
\includegraphics[scale=0.5]{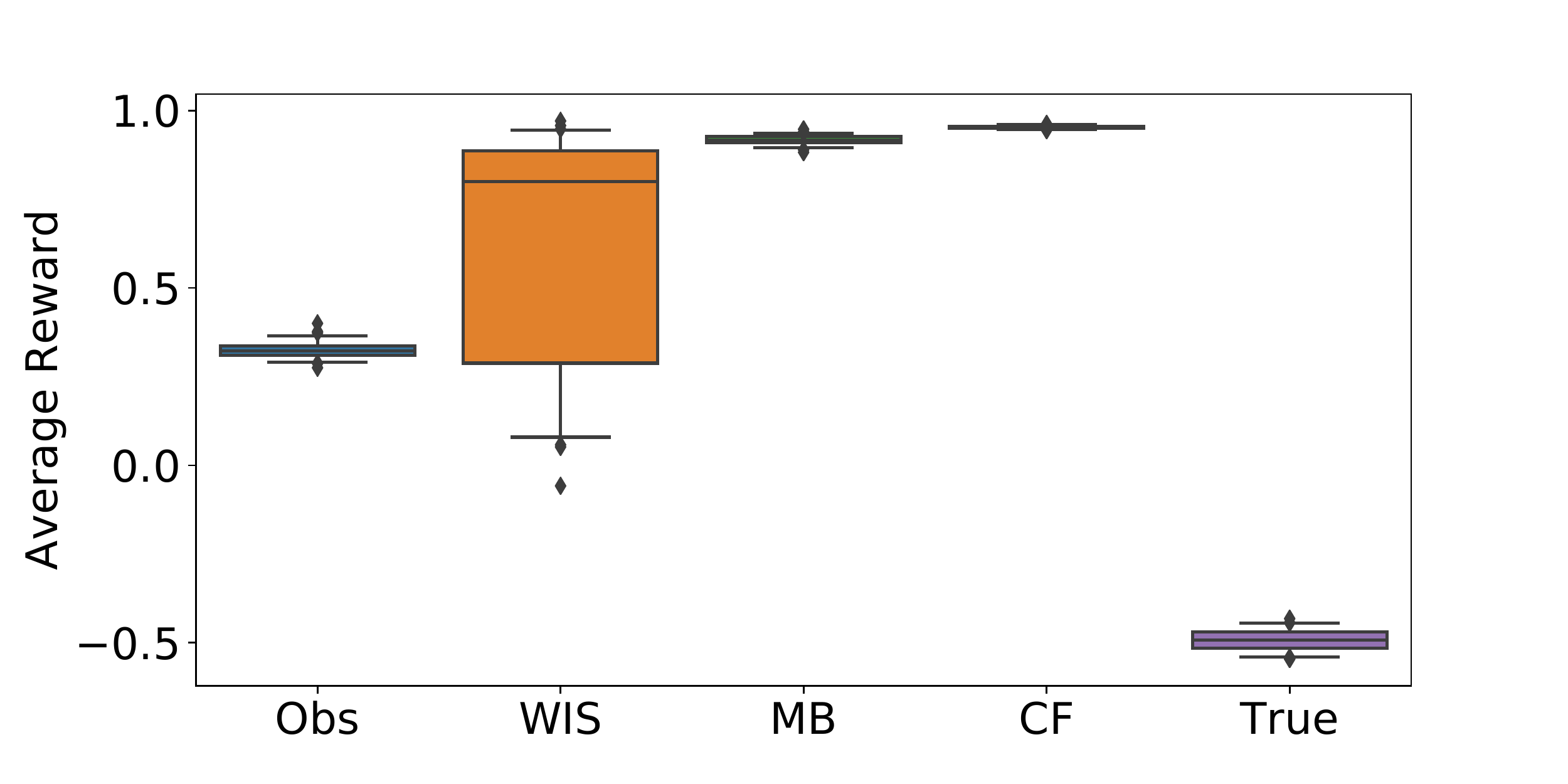}
\caption{Box plots of observed reward under the behavioural policy and the estimated reward for the target policy using WIS, MB and CF policy evaluation methods. The true reward observed for newly simulated trajectories under the target policy is much lower. Figure obtained from \cite{oberst2019}}
\label{fig:evaluation}
\end{figure}

To mimic the situation of partial observability, the glucose and diabetes state are hidden when learning the target policy. The resulting training data is used to learn a model of sepsis using normalized empirical counts of state transitions. Policy iteration is used to learn the target policy, while restricting the action space to the set of observed actions. Under this experimental setup, since the true counterfactual trajectories can be computed using the simulator, the authors reveal how off-policy evaluation and batch reinforcement learning can be optimistic. First, the target policy was evaluated using weighted importance sampling (WIS), and model-based policy evaluation (MB-PE) using the parameters of the learned sepsis model.  As shown in Figure \ref{fig:evaluation}, both methods estimate a high value for the target policy, with WIS having a noticeably high variance. Second, counterfactual policy evaluation (CPE) was performed by sampling counterfactual trajectories from the Gumbel-Max SCM constructed from the learned transition probabilities. CPE also overestimates the value of the learned policy, reflecting the dangers of unobserved confounders. 

\begin{figure}[h!]
\centering
\includegraphics[scale=0.5]{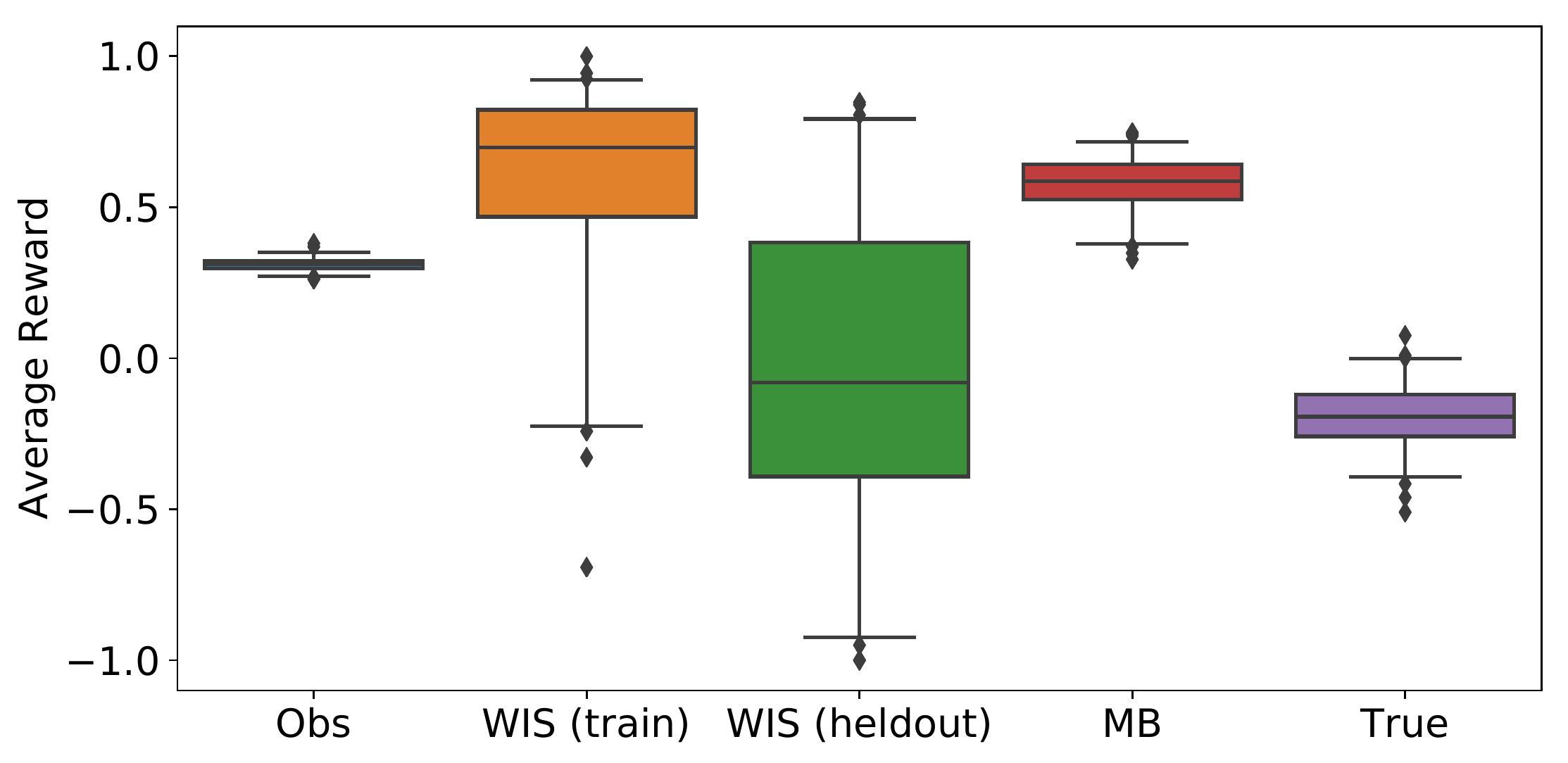}
\caption{Box plots of observed reward under the behavioural policy and the estimated reward for the target policy using WIS on training and heldout data, and MB policy evaluation methods. The target policy was learned with full access to the state. MB policy evaluation overestimates the policy's value.}
\label{fig:ope_obs_results}
\end{figure}

The authors then generated trajectories under the target policy with the sepsis simulator, and observed that the target policy performs poorly. Specific counterfactual trajectories can be examined to understand why off-policy evaluation overestimates the target policy. Using the learned SCM the authors sample counterfactual trajectories for each individual patient in the validation set under the target policy and examine the trajectories of patients whose outcome is predicted to be discharge under the target policy, but whose observed outcome under the clinician's policy is mortality. For instance, the target policy recommended stopping treatment for a specific patient whose vital signs were normal except for glucose levels which were dangerously low. Since diabetes and glucose fluctuations are rare in the simulated data, and since they are not observed, the learned SCM predicts that the patient has a high chance of recovery since the other vital signs are normal. When the full state was observable to the RL algorithm, WIS on heldout data resulted in a median estimated value closer to the true value of the target policy. However model-based (MB) off-policy evaluation continues to overestimate the target policy as shown in Figure \ref{fig:ope_obs_results} which suggests that the learned transitions continue to suffer from extrapolation errors.

While the experiments of Oberst and Sontag  \cite{oberst2019} are rather simplified, they illustrate the susceptibility of reinforcement learning to unobserved confounders and extrapolation errors due to limited observational data. Accounting for the non-randomization of policies found in medical records, and how it influences the learned policy, is a crucial step towards the applicability of reinforcement learning in health care. It is well known that supervised methods for risk stratification are susceptible to confounding factors \cite{paxton2013developing}. For example \cite{Caruana:2015:IMH:2783258.2788613} remarked that since asthmatic patients are more likely to receive intensive care for pneumonia than non-asthmatic patients, mortality predictors can learn that asthmatic patients are at a lower risk of mortality than non-asthmatic patients. Gottesman {\em et al.}~\cite{gottesman2019guidelines, gottesman2018evaluating} reasoned that reinforcement learning is not immune to such confounding. Combining counterfactual models with reinforcement learning could be used to adjust for spurious correlations resulting from the clinicians' policies. 

\section{Validation of Decision Support}

Since the use of decision support tools include a practitioner in-the-loop, such tools do not exempt health practitioners from the responsibility for their decisions. While it is ultimately up to the health practitioner to decide whether to follow the treatment regime suggested by an RL agent, or ignore it if it isn't a reasonable suggestion, this does not mean that machine learning or reinforcement learning models can be treated as a black box. A model that produces a questionable suggestion even once will be difficult to trust subsequently and clinicians may forgo its use altogether, thus nullifying the intended benefit of the decision support tool. 

A commonly used demonstration of the validity of the learned policy is the plot of the frequency of mortality in the dataset against differences between administered dosages and recommended dosages by the learned policy. A characteristic U-shape is often observed, where the frequency of mortality is shown to increase as the difference between the administered and recommended dosages increases. While the mortality rate should be lower when the clinician's and learned policies agree, the higher mortality rate of patients who were administered different dosages than the recommended treatment should not be interpreted as evidence for the validity of the learned policy. Additionally, as shown by Gottesman {\em et al} \cite{gottesman2018evaluating} a similar U-curve can even be obtained by comparing the clinicians' policy with a no treatment or random policy. The U-curve for a no-treatment policy simply shows the correlation between mortality and higher administered dosages of IV and vasopressors. Since the distribution of administered dosages has a long tail, discretizing the action space based on quantiles essentially places most observed actions into the first bin, corresponding to zero-dosage. It is therefore no surprise that the learned policy and the no-action policy result in similar U-curves. 

 Gottesman {\em et al.}~\cite{gottesman2018evaluating} suggested that the learned policy can be sensitive to the choice of state representation. Since off-policy evaluation is performed using the chosen state representation, it can be difficult to ascertain the validity of the learned policy. A suggested action could be supported by showing which parts of the state representation or observation vector most affected the choice of action recommendation. For example, the suggestion to administer antibiotics can be supported by pointing to the white blood cell count or the results of a culture test. Ribeiro {\em et al.}~\cite{ribeiro2016should} provided indicators of feature influence by approximating a deep model locally to a more interpretable model of lower complexity. Similary, Chen {\em et al.}~\cite{Chen2018} developed a feature selector, trained to select features by maximizing the mutual information between the selected features and the model's output. It is worth investigating whether incorporating solutions for interpretability in decision support can  
be effective at revealing how the model is influenced by correlations between the behavioural policy and patient features. 

In the case of discovering better treatment strategies, highlighting the influence of patient features may not be enough to explain a learned policy, since it does not reveal how the feature is interpreted by the agent. When deciding on a course of action for a given patient, a physician can be influenced by non-clinical factors such as the patient's socioeconomic status, age, or gender \cite{hajjaj2010non}. While age and gender could be considered as clinical factors, since they are related to co-morbidities \cite{hajjaj2010non}, they can also have non-clinical influences. Factors such as a patient's comfort and preferences could also lead to unnecessary or sub-optimal interventions, such as unnecessarily prescribing antibiotics due to pressure from patients \cite{hajjaj2010non}. Although the state representation could exclude these non-clinical factors, they may still play a role. E.g., if a patient's comfort is correlated with compliance, then it may be beneficial to consider a patient's preferences when suggesting a course of treatment. Since some features can have a clinical and non-clinical influence on decisions, it can be challenging to understand the nature of their influence on the learned policy. Another challenge arises from the need to limit treatment costs while improving patient outcomes. These objectives could come into conflict, and it might be unclear which objective a chosen action favours. 

\section{Conclusion}

While reinforcement learning provides a natural solution for learning improved policies for sequential tasks,
its application in healthcare introduces difficulties with regards to describing the state and action space, learning and evaluating policies from observational data, and designing reward functions. Since the chosen action and state representations affect the learned policy as well as its estimated value, off-policy evaluation may not be enough to reveal shortcomings of the learned policy. Additionally, when using observational data to learn a model of patient progression, one must be careful about distributional shifts incurred by a change in policy which might invalidate the model in under-explored areas of the state space. A potential area for future research is the development of standards or heuristics to validate and diagnose learned policies. Such methods should be centered around how to account for rare or unseen states, and understanding the effect of the chosen state and action representations. For instance, a continuous state representation requires a parameterized $Q$-function which is more susceptible to extrapolation errors; however, a discretized state space and a tabular $Q$-function may result in a loss of granularity. How should designers interpret differences in the learned policy under different design choices? What kinds of improvement should researchers reasonably expect given the limit of observed state-action pairs? Given the safety-critical nature of recommending treatment actions, validating learned policies will require more than general guidelines if RL is to become a widely adopted approach for personalized treatment. 

\section*{Acknowledgements}
FR is a CIFAR Chair in Artificial Intelligence.

\bibliographystyle{plainnat}
\bibliography{references}

\end{document}